\documentclass[letterpaper, conference]{ieeeconf}
\IEEEoverridecommandlockouts            

\usepackage{cite}
\usepackage{amsmath,amssymb,amsfonts}
\usepackage{algorithmic}
\usepackage{graphicx}
\usepackage{textcomp}
\usepackage{xcolor}
\usepackage{comment}
\usepackage{booktabs}
\usepackage[font=normalsize,labelfont=bf]{caption}

\title{Co-optimizing Physical Reconfiguration Parameters and Controllers for an Origami-inspired Reconfigurable Manipulator}

\author{Zhe Chen$^{1}$, Li Chen$^{2}$, Hao Zhang$^{2}$, and Jianguo Zhao$^{1}$
\thanks{$^{1}$ Zhe Chen and Jianguo Zhao are with the Department of Mechanical Engineering at Colorado State University, Fort Collins, CO, 80523, USA. E-mail: \{Zhe.Chen, Jianguo.Zhao\} @colostate.edu.
}
\thanks{$^{2}$ Li Chen and Hao Zhang are with the Manning College of
Information and Computer Sciences, University of Massachusetts Amherst,
Amherst, MA 01002, USA. Email: \{lchen0, hao.zhang\}@umass.edu.
}
}

\begin{document}
\maketitle

\begin{abstract}
Reconfigurable robots that can change their physical configuration post-fabrication have demonstrate their potential in adapting to different environments or tasks.
However, it is challenging to determine how to optimally adjust reconfigurable parameters for a given task, especially when the controller depends on the robot's configuration.
In this paper, we address this problem using a tendon-driven reconfigurable manipulator composed of multiple serially connected origami-inspired modules as an example. Under tendon actuation, these modules can achieve different shapes and motions, governed by joint stiffnesses (reconfiguration parameters) and the tendon displacements (control inputs). We leverage recent advances in co-optimization of design and control for robotic system to treat reconfiguration parameters as design variables and optimize them using reinforcement learning techniques.
We first establish a forward model based on the minimum potential energy method to predict the shape of the manipulator under tendon actuations. 
Using the forward model as the environment dynamics, we then co-optimize the control policy (on the tendon displacements) and joint stiffnesses of the modules for goal reaching tasks while ensuring collision avoidance. 
Through co-optimization, we obtain optimized joint stiffness and the corresponding optimal control policy to enable the manipulator to accomplish the task that would be infeasible with fixed reconfiguration parameters (i.e., fixed joint stiffness).  
We envision the co-optimization framework can be extended to other reconfigurable robotic systems, enabling them to optimally adapt their configuration and behavior for diverse tasks and environments. 
\end{abstract}


\section{Introduction}

Traditionally, the design and control of robotic systems have been treated as separate processes: a robot’s physical structure is first designed, and then a controller is developed to operate it. 
However, co-design or co-optimization---the simultaneous optimization of both a robot’s physical design and its control strategy---has recently emerged as a new method, spurred by recent advancements in learning-based control, particularly those leveraging simulation-based training for zero-shot deployment \cite{miki2022learning,kumar2021rma}.
This co-optimization approach enables robotic systems to achieve optimal performance by exploring synergies between morphology and behavior.  Notable examples include legged robots \cite{belmonte2022meta,dinev2022versatile}, soft robots \cite{schaff2022soft}, robotic hands \cite{chen2021co}, and modular robots \cite{gupta2022metamorph}, where integrated design and control optimization have led to improvements in efficiency, adaptability, and robustness.

However, most existing work on co-optimization generally focuses on geometric dimensions as design parameters, such as the leg length for legged robots \cite{belmonte2022meta}, which are fixed after fabrication and difficult to modify \cite{nygaard2021real}.
To enable robots capable of adapting their morphology and behavior on the fly to accommodate different tasks or environments, it is crucial to consider parameters that can be adjusted or reconfigured post-fabrication, and we call them physical reconfiguration parameters.
Examples of such reconfiguration parameters include curvatures for body/leg parts \cite{baines2022multi,sun2023embedded} and joint stiffness, which can be actively tuned based on the advancements in variable stiffness materials \cite{buckner2019enhanced}.
Tunable joint stiffness, in particular, enables programmable motion in mechanical systems such as origami \cite{stern2020supervised,lerner2024reconfigurable} and linkage-based mechanisms \cite{sun2019adaptive}, enhancing their adaptability after fabrication.

\begin{figure}
    \centering
    \includegraphics[width=1\linewidth]{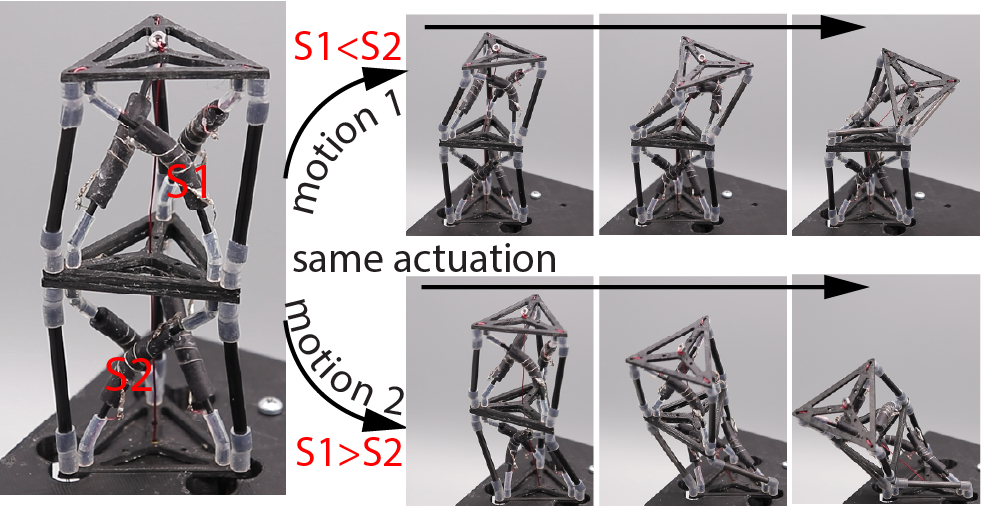}
    \caption{Illustration of programmable motion for two serially connected origami-inspired modules \cite{chen2022origami}. $S_1$ and $S_2$ represent the stiffness of a joint in the top and bottom module, respectively. When $S_1>S1$, the manipulator undergoes motion 1. If $S_1<S_2$, the manipulator undergoes motion 2 with the same actuation. \vspace{-20pt}}
    \label{fig:PrototypeIllustration}
\end{figure}

We have recently demonstrated an origami-inspired reconfigurable module capable of achieving programmable shapes and motions under tendon actuation and different stiffness for selected joints \cite{chen2022origami}. 
Fig. \ref{fig:PrototypeIllustration} shows a manipulator with two serially connected modules. When the stiffness of one joint in the top module $S_1$ is larger than the stiffness of one joint in the bottom module $S_2$, the manipulator undergoes motion 1. If $S_1<S_2$, the manipulator undergoes motion 2 with the same actuation (detailed working principle can be found in \cite{chen2022origami} and section \ref{sec:Origami_inspired_reconfigurable_manipulator}).
With more modules connected in series, the resulting manipulator can achieve more diverse motions, without changing the geometric dimensions, but by tuning the stiffnesses of selected joints within each module.

The contribution of this paper is to leverage co-design framework to jointly optimize physical reconfiguration parameters and controllers for reconfigurable robotic systems. Specifically, we co-optimize the joint stiffnesses and tendon actuations for the reconfigurable origami-inspired manipulator to accomplish desired tasks such as reaching a goal position while avoiding certain objects. 
We address the problem by considering the reconfiguration parameters (i.e., joint stiffness) as design parameters (instead of the traditionally used geometric dimensions) and tendon actuations as control actions. 
Specifically, we maintain a Gaussian distribution over the joint stiffnesses and uses reinforcement learning to optimize both the neural network control policy and the distribution parameters to maximize the expected return of the control policy over the stiffness distribution. 
Co-optimizing the joint stiffnesses and tendon actuations can generate robotic manipulator that can adapt to task requirement post fabrication if we can control the stiffness at a specific value before the tendons are actuated.

The rest of this paper is organized as follows. Related works are discussed in Section \ref{sec:related_work}. After that, we explain the working principle, and develop a forward model of the origami manipulator in Section \ref{sec:Origami_inspired_reconfigurable_manipulator}. We then discuss why co-optimization is necessary for the manipulator in Section \ref{sec:necessity}. After that, we discuss how we implement the co-optimization of the design parameters and control algorithm of the manipulator for a reaching task as well as the results. Lastly, conclusions are drawn in Section \ref{sec:Conclusion}.

\section{Related Work} \label{sec:related_work}

\textbf{Programmable motion with variable stiffness joints.}
For origami robots, the stiffnesses of creases can influence their mechanical properties and dynamic behaviors. Moreover, the ability of origami systems to dynamically adjust their stiffness in real time could greatly expand their applicability in robotic tasks, such as locomotion, manipulation, and grasping. Firouzeh \textit{et al.} \cite{firouzeh2017grasp} used shape memory polymer (SMP) for an underactuated origami gripper, which can change the stiffness by heating up the SMP joints. Lin \textit{et al.} \cite{lin2020controllable} employed laminar jamming to control the stiffness of an origami structure on the fly. Lerner \textit{et al.} \cite{lerner2024reconfigurable} developed a novel variable stiffness joint and demonstrated the programmable motion of an origami robot with those joints. 

\textbf{Model-based co-optimization.} Some researchers used detailed dynamic models of the robots for model-based co-optimization. For instance, Spielberg \textit{et al.} \cite{spielberg2017functional} demonstrated that robot design parameters can be incorporated into trajectory
optimization process, enabling the concurrent optimization of robot trajectories and physical designs. Deimel \textit{et al.} \cite{deimel2017automated} used a simplified model for the dynamics of the soft robotic hand in the co-design process, which updates the design parameters with particle filter optimization method.
Liao \textit{et al.} \cite{liao2019data} proposed hierarchical process constrained batch Bayesian optimization (HPC-BBO) to automatically optimize robot morphologies and the controllers in a data efficient manner. 

\textbf{Data-driven co-optimization.} Data-driven approaches, such as deep reinforcement learning (RL), have proven highly effective in addressing the complex dynamics of robotics and their interactions with the environment \cite{nagabandi2020deep}. Compared with model-based co-optimization methods, RL-based co-optimization methods excel in learning directly from interactions with the environment.
Recently, researchers have also explored implementing co-optimization using RL methods \cite{yuan2021transform2act, he2024morph, islam2024task, spielberg2019learning}. For instance, He \textit{et al.} \cite{he2024morph} proposed the MORPH framework to co-optimize robot morphology and control policy using a neural network based proxy model to approximate the real physical model of the robot. Wang \textit{et al.} \cite{wang2019neural} proposed Neural Graph Evolution to co-evolve both the robot design and the control policy, representing a robot's morphology with a graph neural network. Chen \textit{et al.} \cite{chen2023deep} developed a bi-level optimization method to co-optimize both morphology parameters and control policy for small-scale legged robots.

\begin{figure}[]
    \centering
    \includegraphics[width=\linewidth]{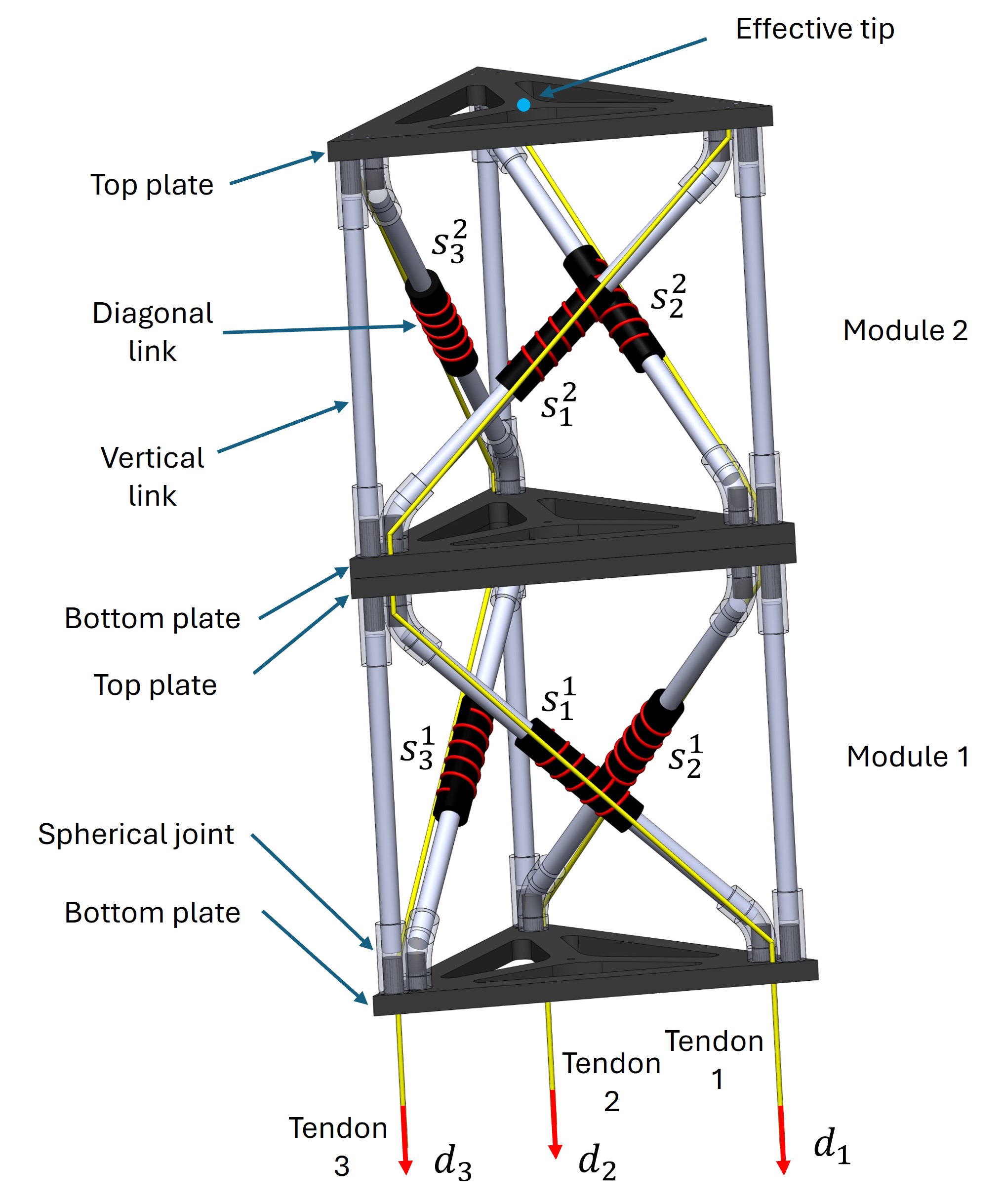}
    \caption{A reconfigurable manipulator consisting of two origami-inspired modules connected in series}
    \vspace{-10pt}
    \label{fig:solid_model_manipulator}
\end{figure}

\section{Origami-inspired reconfigurable manipulator} \label{sec:Origami_inspired_reconfigurable_manipulator}

In this section, we discuss the working principle of the reconfigurable manipulator and the forward kinematic model that can predict the manipulator's motion given joint stiffness and tendon actuation.

\subsection{Working Principle}

A manipulator consisting of two origami-inspired reconfigurable modules connected in series is shown in Fig. \ref{fig:solid_model_manipulator}. We refer to the top module as Module 2 and the bottom one as Module 1. Each module has a top and a bottom triangular plate that are connected by three pairs of vertical and diagonal links. These links are attached to the plates through silicone tubes, which function as compliant spherical joints. The bottom plate of Module 1 is fixed. On each side of the triangular manipulator, an actuation tendon (shown in yellow in Fig. \ref{fig:solid_model_manipulator}) is anchored to the top plate of Module 2, routed along the diagonal links and threaded through the plates in a zigzag pattern. The tendons extend through the fixed bottom plate of Module 1 and are ultimately connected to motors placed beneath the bottom plate, which control the displacements of the tendons. 

Each diagonal link is implemented as a \underline{V}ariable \underline{S}tiffness \underline{J}oint (VSJ), consisting of a thermoplastic material enclosed by an elastic tube in the middle. We can reconfigure the stiffness of each VSJ through Joule heating by using heating wires around the tube. The detailed working principle for the VSJs is introduced in our earlier work \cite{chen2022origami,sun2019adaptive}. Depending on the SVJs' stiffness values, each module exhibits distinct motion characteristics in response to tendon actuation. By connecting multiple modules in series, the manipulator can achieve more complex and versatile motions, significantly expanding its workspace and functional capabilities. For a manipulator, we define the centroid of the top plate of the topmost module as its effective tip.

\subsection{Forward Model} \label{sec:forward_model}
To predict the shape of the origami manipulator, which consists of \( N \) serially connected modules, a forward model is required to determine its deformation given the displacements of the three tendons, represented as \( D = [d_1, d_2, d_3]^T \), and the stiffnesses of all joints, represented as $S=[s_1^1,s_2^1,s_3^1,s_1^2,s_2^2,s_3^2,...,s_1^N,s_2^N,s_3^N]^T$.
Unlike our previous design in \cite{chen2022origami}, the current manipulator is driven by three tendons, necessitating the development of a new forward model for the co-optimization process. To achieve this, we employ the minimum potential energy method, which determines the equilibrium shape of the manipulator by considering both the applied tendon displacements and the stiffnesses of all VSJs.

\begin{figure}[]
    \centering
    \includegraphics[width=\linewidth]{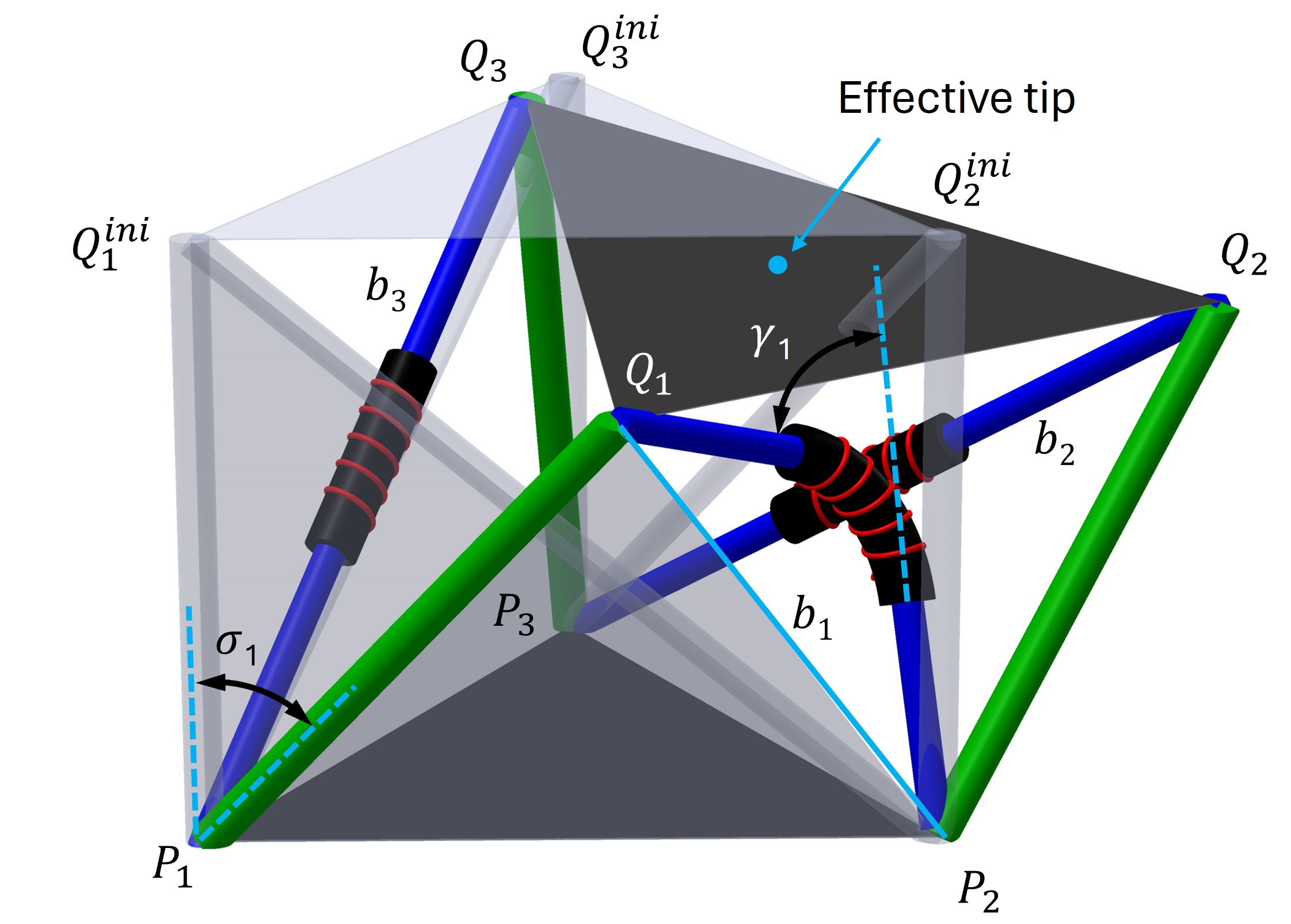}
    \caption{Geometry of the deformed module}
    \label{fig:forward_model}
\end{figure}

We first illustrate important parameters for the forward model. In Fig. \ref{fig:forward_model}, the initial shape of Module 1 in Fig. \ref{fig:solid_model_manipulator} is shown in transparent. To simplify the model, we assume that every neighboring pair of vertical link and diagonal link is connected to the top or bottom plate at the same point, meaning they share the same spherical joint. For instance, vertical link $P_1 Q_1^{ini}$ and diagonal link $P_2 Q_1^{ini}$ are connected to the top plate $Q_1^{ini}Q_2^{ini}Q_3^{ini}$ at the same point $Q_1^{ini}$. If link $P_2Q_1$ is soft and $P_3Q_2$, $P_1Q_3$ are rigid, the module under actuation would deform to a shape shown in nontransparent in Fig. \ref{fig:forward_model}.

For a manipulator made from $N$ modules, the shape of $j$-th module is uniquely represented by the chord lengths of its three diagonal links $[b_1^j,b_2^j,b_3^j]^T$. Note that if a diagonal link is straight, the chord length equals its initial length $b_{ini}$. In this case, the shape of the manipulator can be represented as $B=[b_1^1,b_2^1,b_3^1,b_1^2,b_2^2,b_3^2,...,b_1^N,b_2^N,b_3^N]^T$.
Since the tendons cannot extend, the manipulator is subject to the following constraint equation.
\begin {equation}
\label{eqn:constraint_tendon}
d_i \leq N \cdot b_{ini} - \sum_{j=1}^{N} b_i^j, \quad i = 1,2,3
\end {equation}
where $i$ corresponds to the $i$-th side of the manipulator. 

However, it is possible that infinite many sets of $B$ may satisfy the same tendon constraint. We use the minimum potential energy method to choose the set of $B$ from the infinite many possible $B$s that minimizes the potential energy of the manipulator, $E$. For each module, the potential energy is calculated as follows.

\begin{equation}
\label{eqn:energy}
 E^j =\frac{1}{2} \sum_{i=1}^{3} s_i^j (\gamma_i^j)^2 + \frac{1}{2} \sum_{m=1}^{12} k_m^j(\sigma_m^j)^2
\end{equation}
where the first item represents the potential energy stored in the three VSJs, the second item represents the potential energy stored in the elastic spherical joints connecting the links and the plates. $\gamma_i$ denotes the bending angle of the VSJs. $\sigma_m$ denotes the bending angle of the spherical joints. $s_i$ and $k_m$ are the corresponding effective stiffnesses of the VSJs and the spherical joints. $\gamma_1$ and $\sigma_1$ are shown in Fig. \ref{fig:forward_model}. Note that $\gamma_i$ and $\sigma_m$ can be obtained from the shape of the manipulator, $B$, through geometric calculations \cite{chen2022origami}.

To model real-world constraints, we impose a limit on the maximum force each cable can generate, as a physical prototype would rely on motors with finite stall torque. In this way, the manipulator is also subject to the following constraint equation of force limit, $F_l$. 

\begin {equation}
\label{eqn:constraint_tendon_force}
\frac{\partial E}{\partial d_i} \leq F_l, \quad i = 1,2,3
\end {equation}

We formulate the forward model as an optimization problem as follows
\begin{equation} \label{eqn:optimization_equation}
\begin{aligned}
    \min_{B} \quad & E = \sum_{j=1}^{N} E^j \\
    \text{s.t.} \quad & \text{Constraints (Eq. \ref{eqn:constraint_tendon}, Eq. \ref{eqn:constraint_tendon_force})}
\end{aligned}
\end{equation}

The solution $B$ of the optimization problem depends on both the stiffnesses of the VSJs $S$ and the tendon displacement $D$, since $S$ and $D$ are included in the objective function and constraint equations in Eq. \eqref{eqn:optimization_equation}.
With an optimal $B$, we can solve the forward problem to predict the final shape of the manipulator consisting of multiple modules connected in series, given $D$ and $S$. We note that though we could only achieve binary stiffness of the VSJs (either rigid or soft) in our earlier work \cite{chen2022origami}, the forward model presented here assumes continuously adjustable stiffness. This enhancement makes the model more general and applicable to a wider range of scenarios. The position of the effective tip of a module or a manipulator can be readily calculated based on the shape of the module or the manipulator.

\section{Necessity of Co-Optimization for joint stiffness and tendon actuation}\label{sec:necessity}
In this section, we discuss why co-optimization is needed. 
Specifically, we first show that with different stiffnesses of the VSJs, the motion of the manipulator can be different under the same tendon actuation. 
We then show that the reachable workspace can also be different under different set of stiffness values.

\subsection{The same actuation can generate different trajectories under different VSJ stiffnesses}

\begin{figure}[]
    \centering
    \includegraphics[width=0.9\linewidth]{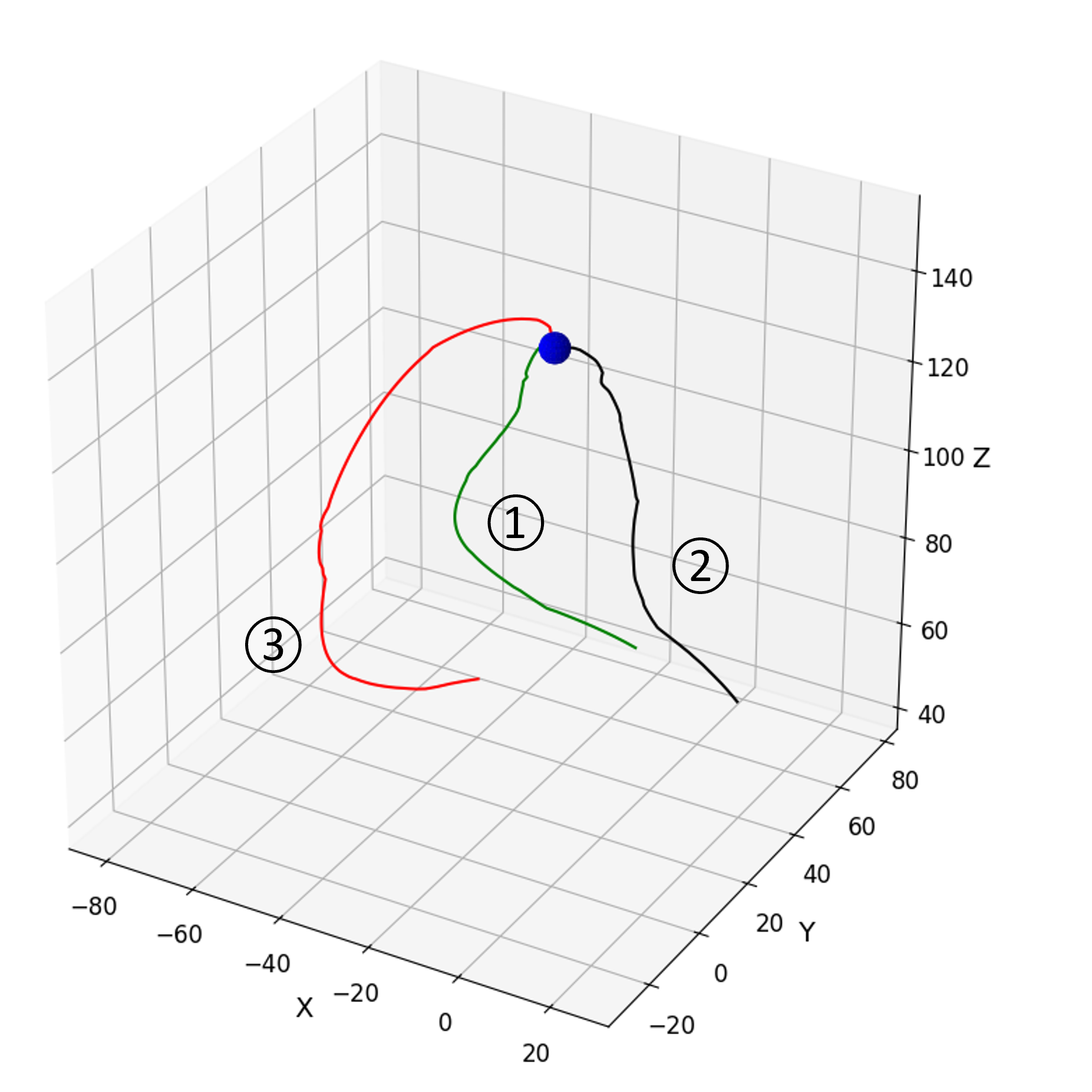}
    \caption{The manipulator under fixed actuation sequence exhibits different motions if the VSJs have different stiffnesses. Trajectories 1, 2, and 3 correspond to stiffness set $S_1$, $S_2$, $S_3$, respectively.}
    \vspace{-20pt}
    \label{fig:different_trajectories}
\end{figure}

To illustrate how the motion depends on the stiffness, we use a manipulator made from four origami modules and predict its motion using the forward model developed in section \ref{sec:forward_model} with the following actuation for the displacements of the three tendons.
\begin{equation}
    \Delta_1 = 128t,\quad \Delta_2 = 100\sin{\frac{\pi}{2}t}, \quad \Delta_3 = 100t^2
\end{equation}
where the parameter $t$ is in the range $(0,1)$.
We choose three different sets of stiffnesses as follows. $S_1$ = [1.37, 1.65, 0.66, 1.30, 0.64, 0.97, 0.68, 2.28, 0.76, 0.42, 1.23, 1.06]$^T$ is randomly sampled from a uniform distribution within the range \( (0.4, 2.5) \). $S_2$ is obtained by shifting $S_1$ by 1 unit to the right, and $S_3$ is generated by shifting $S_1$ by 2 units to the right. 
The same actuation results in drastically different trajectories for the position of the effective tip, as shown in Fig. \ref{fig:different_trajectories}. The blue dot represents the initial position of the effective tip when the manipulator is not actuated.


\subsection{Different stiffness can lead to different workspace} 

\begin{figure}[]
    \centering
    \includegraphics[width=0.8\linewidth]{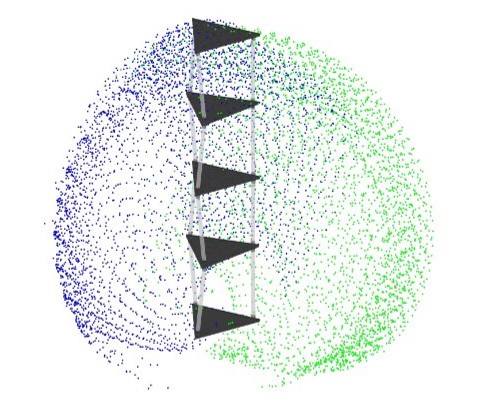}
    \caption{The manipulators with different stiffness selections have varying reachable workspace.}
    \vspace{-15pt}
    \label{fig:different_workspace}
\end{figure}

We further demonstrate that the same manipulator, with different VSJ stiffnesses, can achieve varying reachable workspaces. 
We obtain the workspace by uniformly sampling tendon displacements within their feasible ranges and computing the corresponding end-effector position for each sample. The workspace varies with the stiffness selections since the force limit is included in the forward model. For two different stiffness sets, $S_4$ = [2.50, 0.80, 0.80, 2.50, 0.80, 0.80, 2.50, 0.80, 0.80, 2.50, 0.80, 0.80]$^T$ and $S_5$ = [0.80, 2.50, 0.80, 0.80, 2.50, 0.80, 0.80, 2.50, 0.80, 2.50, 0.80, 0.80]$^T$, we compute and visualize the corresponding workspaces in Fig. \ref{fig:different_workspace}, with green dots representing $S_4$ and blue dots representing $S_5$. 

\section{Co-optimization of VSJ stiffness and tendon actuation} \label{sec:co_optimization}
In this section, we describe our approach to co-optimization by formulating it as a reinforcement learning (RL) problem. We begin with a brief review of RL fundamentals before presenting the formulation of the co-optimization procedure.

\subsection{Reinforcement learning}
The RL problem is generally formulated as a Markov Decision Process (MDP). An MDP can be represented by a tuple (\(\mathcal{S}, \mathcal{A}, \mathcal{F}, r\)), where \( \mathcal{S}\) is the state space, \(\mathcal{A}\) is the action space, \(\mathcal{F}\) is the state transition model, \( r \) is the reward function.
An agent in state \( s_t \in \mathcal{S} \) at time \( t \) takes action \( a_t \in \mathcal{A} \) according to some policy $\pi_\theta$, and the environment returns the agent's new state \( s_{t+1} \in \mathcal{S} \) according to the state transition model \( \mathcal{F}(s_{t+1} | s_t,a_t) \), along with the associated reward \( r_t= r(s_t, a_t) \). The goal is to learn the optimal control policy $\pi_\theta^* : \mathcal{S} \to \mathcal{A}$ mapping states to actions that maximizes the expected return \[ J(\pi_\theta) = \mathbb{E}_{\tau \sim \pi_\theta} \left[ R(\tau) \right]
\] where $\tau$ is a trajectory obtained by letting the agent act in the environment using the policy $\pi_\theta$,  $R(\tau) = \sum_{i=0}^T \gamma^i r_{t+i}$ is the return for the trajectory $\tau$, where $T$ is the length of the trajectory, $\gamma \in [0, 1)$ is the discount factor of the future rewards.

A stochastic policy, denoted as $\pi_\theta(a_t \mid s_t)$, is commonly used to predict the action $a_t$ given the current state $s_t$. The stochastic nature of the policy  encourages exploration during the training process, enhancing the model's ability to discover optimal actions. Typically, $\pi_\theta(a_t \mid s_t)$ can be modeled as a neural network, which takes the current state $s_t$ as input, and outputs a probability distribution for sampling the action  $a_t$. In most cases, a Gaussian distribution is used for the probability distribution, where the neural network outputs the mean and standard deviation for each dimension of the action $a_t$.  

In this work, we use the Proximal Policy Optimization (PPO) \cite{schulman2017proximal} algorithm to learn an optimal control policy for our reconfigurable manipulator to achieve specific tasks (e.g., reaching goal points).
PPO offers significant advantages over traditional policy gradient methods by providing a stable and efficient training process.
PPO is an on-policy algorithm that alternates between sampling data from the environment and optimizing the following objective:
\begin{equation} \label{eq:PPO formula}
\hat{\mathbb{E}}_t \left[ \min \left( r_t(\theta) \hat{A}_t, \text{clip}(r_t(\theta), 1 - \epsilon, 1 + \epsilon) \hat{A}_t \right) \right]
\end{equation}
where $r_t(\theta) = \frac{\pi_\theta(a_t | s_t)}{\pi_{\theta_\text{old}}(a_t | s_t)}$ is the ratio function, $\hat{A}$ is the advantage function, \(\epsilon\) is the clip range, a small hyperparameter which roughly says how far away the new policy is allowed to go from the old one. We set the clip range \(\epsilon\) to be 0.2. The clipped objective has the effect of maximizing expected return by making only small steps in policy space at a time.

\subsection{Co-optimization}

\begin{figure}[t]  
\centering
\includegraphics[width=\linewidth]{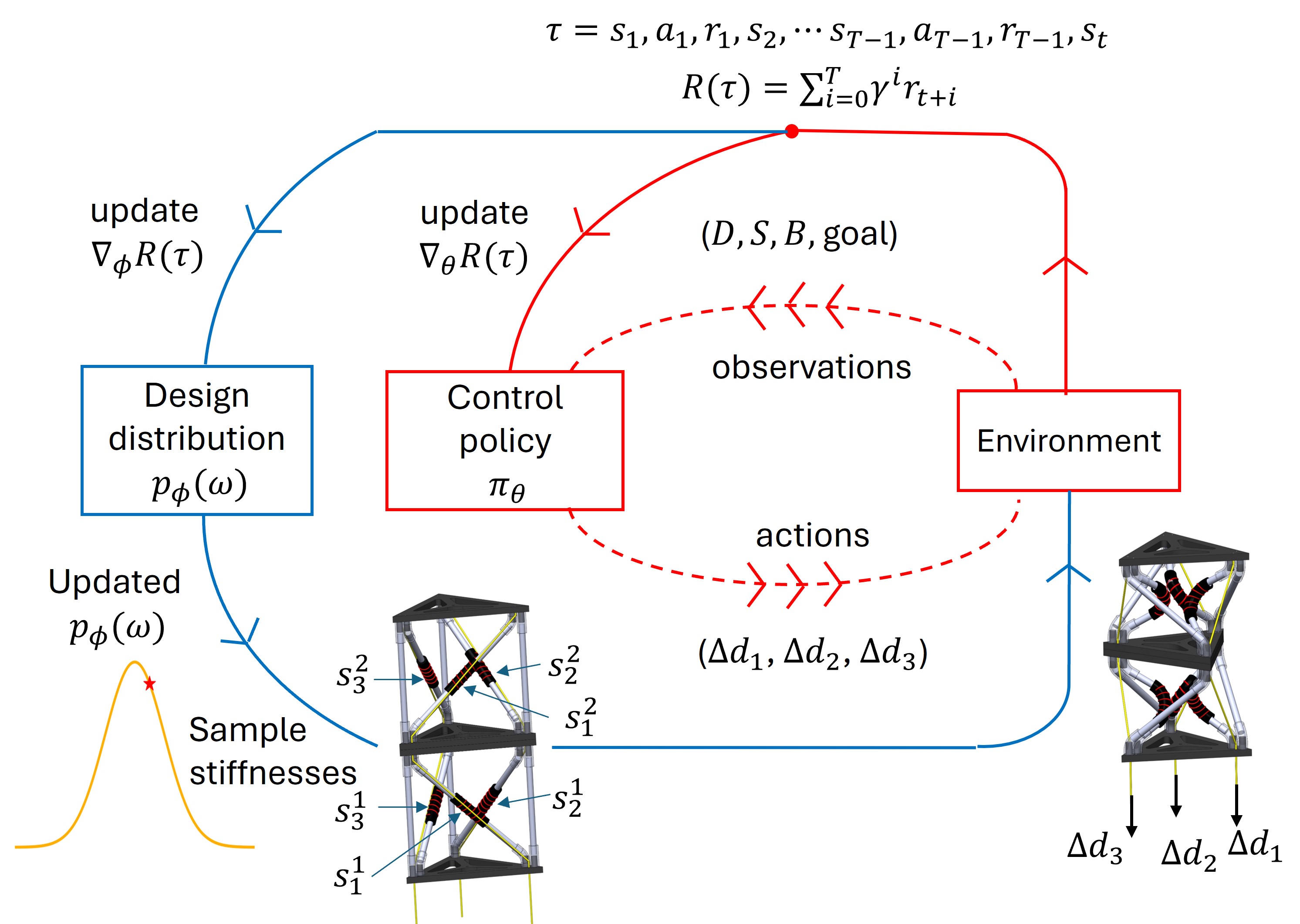}
    \caption{Overview of the co-optimization approach}
    \label{fig:overview of the co-optimization}
\end{figure}

To conduct the co-optimization, we extend the standard RL formulation to include the physical reconfiguration parameters (i.e., the stiffness for VSJs) as a learnable parameter.
Specifically, we co-optimize the reconfiguration parameters and control policy of the robots using the algorithm presented in \cite{schaff2019jointly}, which is inspired by the parameter exploring strategy presented in \cite{sehnke2010parameter}. This algorithm is straightforward to implement and efficient in learning, as it optimizes reconfiguration parameters directly, without the need for an additional neural network or modifying the architecture of the neural network, as used in \cite{he2024morph, yuan2021transform2act}. We denote the design parameters as \( \omega \). For the manipulator, \( \omega \) represents the VSJs' stiffness values.
The goal of our co-optimization is to obtain optimal $\omega^* \in \Omega$ from the space of feasible reconfiguration parameters $\Omega$ that maximizes the agent’s success when used in conjunction with a corresponding optimal control policy $\pi^*_\theta$.

Instead of treating \( \omega \) as fixed values, we model them as Gaussian distributions \( p_\phi(\omega) \) with means $\mu$ and standard deviations $\sigma$. 

\begin{equation}
p(\omega_i) = \frac{1}{\sqrt{2 \pi \sigma_i^2}} \exp\left(-\frac{(\omega_i - \mu_i)^2}{2 \sigma_i^2}\right),
\end{equation}

Now with the Gaussian distribution, we can include the optimization of the design parameters to the classic policy gradient procedure. Similarly to the update of policy parameters, we update the learning parameters $\phi$ of the design distribution ($\mu$ and $\sigma$)  using the episode return. The overview of the co-optimization approach is depicted in Fig. \ref{fig:overview of the co-optimization}, where the red part represents the classic policy gradient procedure, and the blue part represents the optimization of the reconfiguration parameters. The policy function $\pi_\theta(a_t | s_t, \omega)$ now depends on both the current state $s_t$ and  the reconfiguration parameters $\omega$. 
Formally, we seek to find the optimal parameters for both the design and policy  $\phi^*$ and $\theta^*$ such that they can generate maximized expected return:
\begin{equation}
\phi^*, \theta^* = \arg \max_{\phi, \theta} \mathbb{E}_{\omega \sim p_\phi} \left[ \mathbb{E}_{\pi_\theta}[R_\tau] \right].
\end{equation}
At each iteration of training, the policy is trained using PPO to maximize the expected return over designs sampled from the current design distribution $p_\phi$. Also, the design distribution is updated at every iteration to increase the probability density around designs that perform well when using the current learned policy $\pi_\theta$ \cite{schaff2019jointly}:
\begin{equation}  \label{eqn:co_optimization_formula}
\nabla \mathbb{E}_{\omega \sim p_\phi} \left[ \mathbb{E}_{\pi_\theta}[R_t] \right] = \mathbb{E}_{\omega \sim p_\phi} \left[ \nabla \log p_\phi(\omega) \mathbb{E}_{\pi_\theta}[R_t] \right].
\end{equation} 

This shifts the means and standard deviations of the reconfiguration distribution $\phi$ to maximize the expected return under the current policy $\pi_\theta$. 
The term $\nabla \log p_\phi(\omega)$ in Eq. \eqref{eqn:co_optimization_formula} is calculated as follows \cite{sehnke2010parameter}.
\begin{align}
\nabla_{\mu_i} \log p(\omega_i) &= \frac{\omega_i - \mu_i}{\sigma_i^2}, \\
\nabla_{\sigma_i} \log p(\omega_i) &= \frac{(\omega_i - \mu_i)^2 - \sigma_i^2}{\sigma_i^3}.
\end{align}
After the training process, we choose the modes of the reconfiguration distributions as the final VSJs' stiffness values.

\section{Results}

\begin{figure}[]
    \centering
    \includegraphics[width=0.75\linewidth]{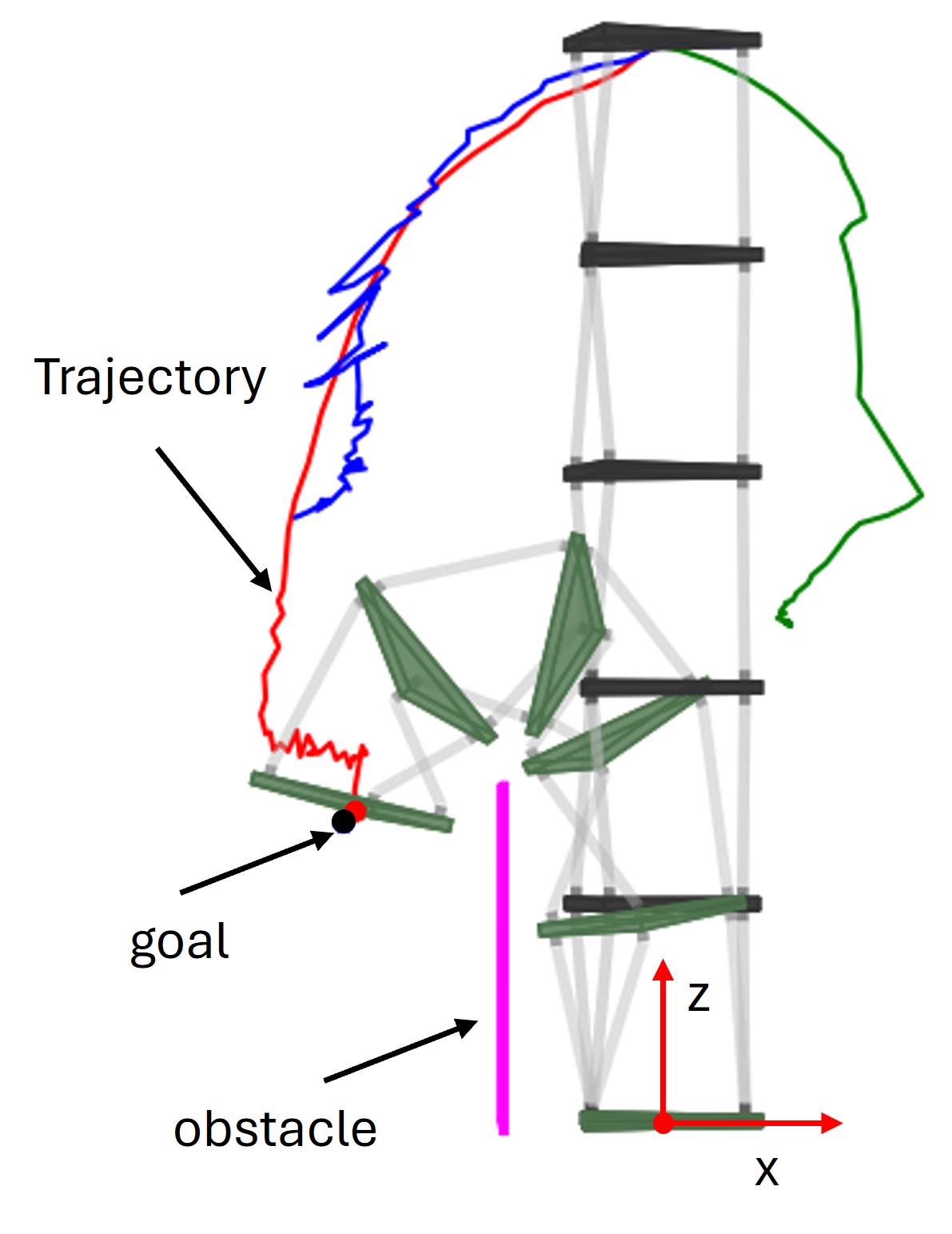}
    \caption{Execution of the trained policy for the reaching task with obstacle avoidance. The red curve shows the trajectory of the effective tip. The red dot shows the position of the effective tip at the last step.}
    \label{fig:single_obstacle}
\end{figure}

In this section, we describe the simulation setup and the corresponding results. 
The forward model in Section \ref{sec:forward_model} is implemented as the state transition model in the learning process. As shown in Fig. \ref{fig:single_obstacle}, we increase the number of modules in the manipulator from 4 to 5 to achieve a more challenging reaching task with obstacle avoidance. The bottom plate of its bottommost module is fixed. We aim to optimize both the stiffnesses of all VSJs and the control strategy on the tendon displacements to make the effective tip of the manipulator reach a goal point while avoiding collision with a surrounding obstacle. 

\begin{figure}[]
    \centering
    \includegraphics[width=1\linewidth]{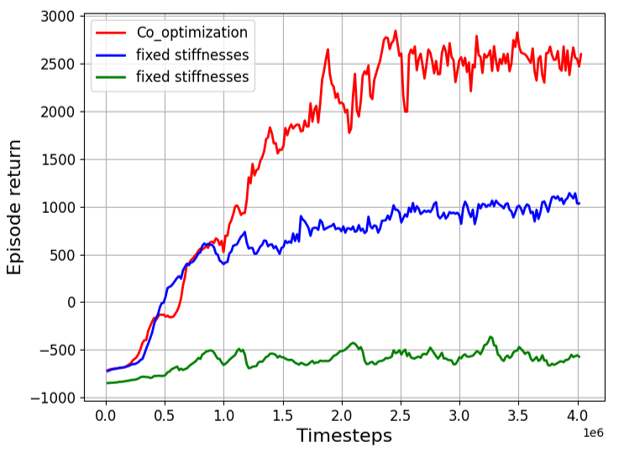}
    \caption{Average return of the training process with one obstacle for a total of 4 million time steps.}
    \vspace{-15pt}
    \label{fig:single_obstacle_curve}
\end{figure}

\subsection{Reaching task with one obstacle}

The objective of this task is to train the RL agent to reach a predefined goal position at \([-50, 0, 50]^T\) while avoiding a planar obstacle. The obstacle is positioned parallel to the YZ plane at a fixed x-coordinate of -25 mm. The agent must navigate through the environment while adhering to kinematic constraints and ensuring collision-free movement.  Since the obstacle lies between the manipulator's initial position and the goal, the manipulator must maneuver around the obstacle to successfully reach the target.
To guide the agent toward the goal position, we employ a dense reward function defined as follows.
\begin{equation}
r = -r_c \mathbf{1}(\text{collision}) + r_s \mathbf{1}(d < d_s) + \frac{r_d}{0.2 + d} 
\end{equation}
where $r_c$ is the collision penalty applied if the agent collides with the obstacle, $r_s$ is the success bonus reward applied if the agent reaches the goal within threshold $d_s$ of 3.0 mm. The indicator function $\mathbf{1}(\text{collision})$ returns 1 if a collision is detected and 0 otherwise. Similarly, the indicator function $\mathbf{1}(d < d_s)$ returns 1 if $d$ is less than $d_s$, and 0 otherwise.
$d= \| p_{t} - p_g \|$ is the distance between the effective tip $p_{t}$ and the goal position $p_g$. The last term represents the distance reward scaled by a constant $r_d$. We extended the forward model to calculate the distance $d$ and to detect the collision between the manipulator and the obstacle.

We employ the PPO algorithm with a multi-layer perceptron (MLP) policy, consisting of two hidden layers with 64 neurons each and ReLU activation. The observation space of the control policy consists of tendon displacements, stiffnesses of the joints, shape of the manipulator, and the goal position. The actions are the change of tendon displacements. The forward model is used to develop a custom environment compatible using the Gymnasium API \cite{towers2024gymnasium} for the RL process. Our co-optimization framework is implemented with the reinforcement learning library Stable-Baselines3 \cite{stable-baselines3}.

To balance exploration and exploitation, we employ a linear decay strategy for both the entropy coefficient and the learning rate. The entropy coefficient decreases linearly from 0.02 to 0.001, while the learning rate gradually declines from 0.00025 to 0. We apply normalization to standardize both observations and rewards to help mitigate large fluctuations and improves learning stability. We present the average episode return as a function of time steps in Fig. \ref{fig:single_obstacle_curve}. 

The stiffnesses of the VSJs are optimized through the training process to be $S$ = [2.50, 2.41, 1.87, 0.58, 2.12, 1.52, 2.50, 0.40, 0.40, 1.26, 0.40, 1.09, 0.82, 1.55, 1.25]$^T$.
To evaluate the trained control policy and stiffness values, we deploy them on a simulated manipulator and visualize the results in Fig. \ref{fig:single_obstacle}. The initial shape of the manipulator is depicted in black and white. The trajectory of the effective tip (shown as the red curve) and the final shape of the manipulator, shown in green, show the agent’s ability to successfully reach the goal (the black dot) while effectively avoiding the surrounding obstacle. We also train control policies for the same reaching task with obstacle avoidance without the co-optimization procedure. In this case, the stiffness values are predetermined as stiffness sets $S_1$ and $S_4$ and remain fixed throughout the learning process. The trained control policies are then deployed on the manipulator with these specified stiffnesses, and the resulting trajectories are shown as the blue and green curves, respectively.
For stiffness set $S_1$, the agent navigates toward the goal but fails to reach it completely. In contrast, for stiffness $S_4$, the agent becomes trapped in a local minimum, avoiding both the goal position and the obstacle.

\subsection{Reaching task with two obstacles}

\begin{figure}[]
    \centering
    \includegraphics[width=0.85\linewidth]{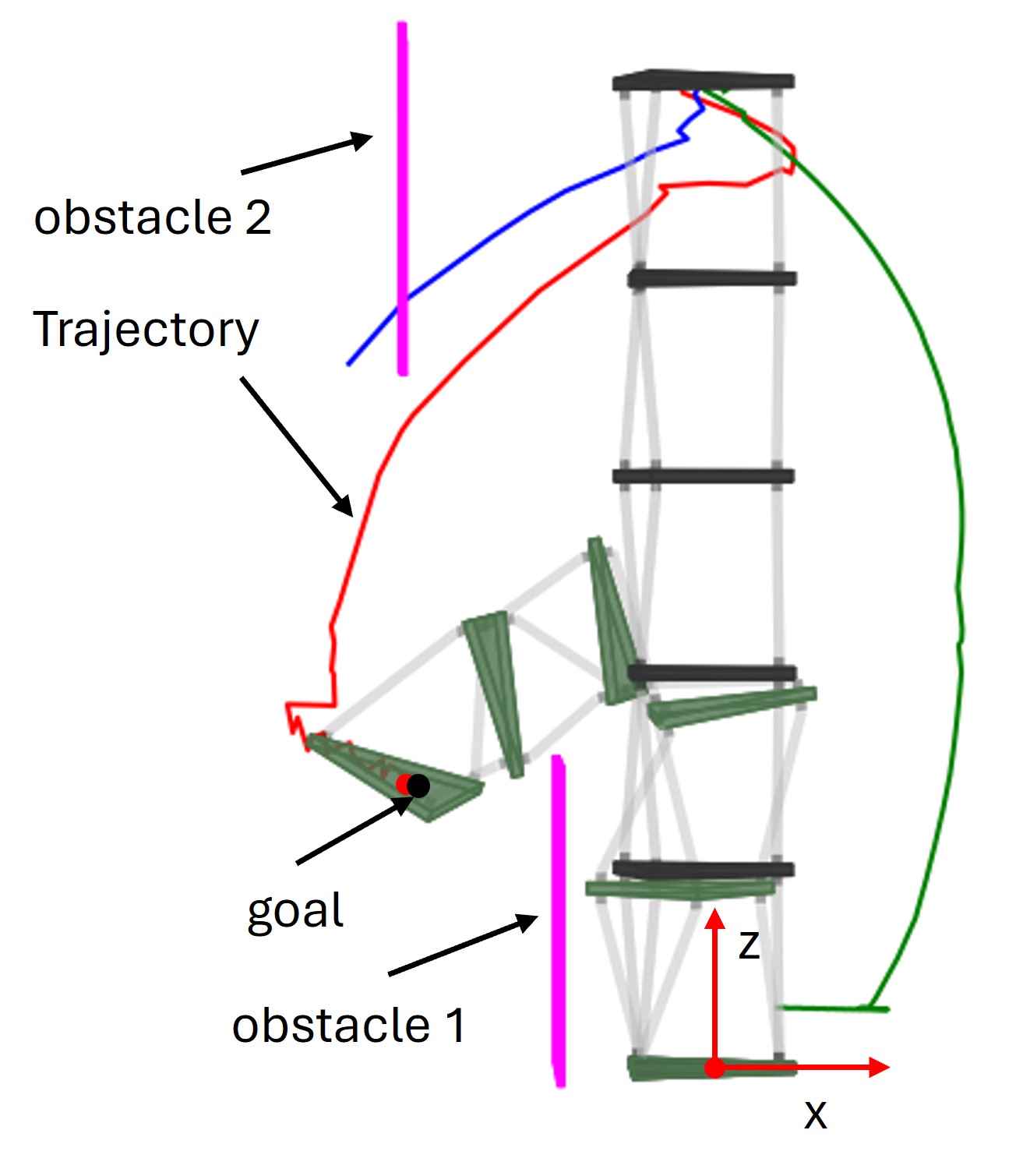}
    \caption{Execution of the trained policy for the reaching task while avoiding two obstacles. The red curve shows the trajectory of the effective tip. The red dot shows the position of the effective tip at the last step.}
    \label{fig:two_obstacles}
\end{figure}

\begin{figure}[]
    \centering
    \includegraphics[width=1\linewidth]{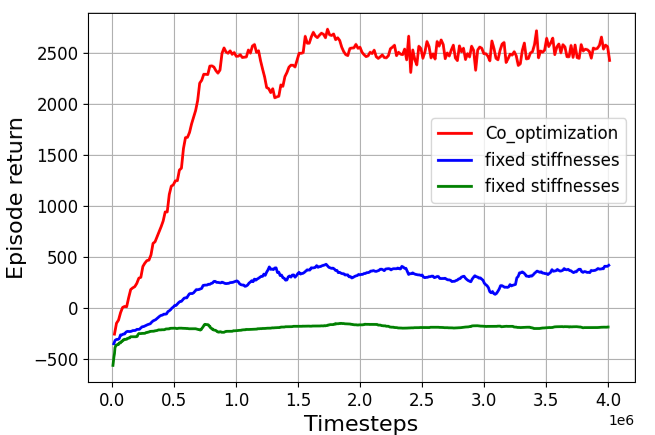}
    \caption{Average return of the training process with two obstacles for a total of 2 million timesteps}
    \vspace{-15pt}
    \label{fig:two_obstacle_curve}
\end{figure}

We increase the task difficulty by introducing a second obstacle (obstacle 2 in Fig. \ref{fig:two_obstacles}), which is parallel to the YZ plane and positioned at a fixed x-coordinate of \(-60\) mm. This obstacle extends from a z-coordinate of 135 mm (bottom edge) to 200 mm (top edge). 

As shown in Fig. \ref{fig:single_obstacle}, the previously trained policy leads the agent to collide with this second obstacle during its motion. 
Unlike the first obstacle, which primarily constrains the final stage of the motion, the second obstacle introduces constraints early in the trajectory, requiring the agent to plan its movement from the very beginning.

Now we include the second obstacle in the custom environment and train a new policy to make the agent reach the same goal point while avoiding both obstacles. The reward function and hyperparameters are the same as before, except that we increase the initial value of the entropy coefficient to 0.03 to encourage greater exploration. 

We present the average episode return as a function of time steps in Fig. \ref{fig:two_obstacle_curve}. The stiffnesses of the VSJs are optimized through the training process to be $S$ = [2.37, 1.70, 2.50, 1.86, 2.04, 2.11, 0.49, 0.82, 1.98, 0.50, 2.50, 2.31, 0.69, 2.50, 0.90]$^T$.
We also deploy the learned control policy and stiffness values on a simulated manipulator and visualize the results in Fig. \ref{fig:two_obstacles}. The resulting trajectory of the effective tip (shown as the red curve) demonstrates that the agent successfully reaches the goal (depicted as the black dot) while effectively avoiding both obstacles. Notably, we observe that the manipulator initially bends slightly to the right before redirecting its motion toward the left to reach the goal. This initial rightward movement allows the manipulator to navigate around obstacle 2 before proceeding toward the target, illustrating a strategic adjustment that ensures collision-free motion. For comparison, we also train control policies for the same reaching task with both obstacle avoidance without the co-optimization procedure. As in the previous case, the stiffness values are predetermined to be $S_1$ and $S_4$ and remain fixed throughout the learning process. The resulting trajectories are shown as the blue and green curves, respectively in Fig. \ref{fig:two_obstacle_curve}.
For stiffness set $S_1$, the agent navigates toward the goal but fails to avoid obstacle 2. For stiffness set $S_4$, the agent again get trapped in a local minimum, avoiding both the goal position and the planar obstacle.

\section{Conclusion} \label{sec:Conclusion}

In this work, we applied a RL-based co-optimization algorithm to jointly optimize the joint stiffnesses and tendon actuation of an origami-inspired reconfigurable manipulator. We first introduced the working principle and developed a forward model for the manipulator, which consists of multiple serially connected origami-inspired modules. We then demonstrated that the manipulator's design parameters, specifically the joint stiffnesses, significantly influence its motion and workspace. Finally, by integrating stiffness optimization into the control learning process, we showed that the co-optimized manipulator outperforms agents with fixed design parameters in reaching tasks while avoiding obstacles. These results underscore the importance of co-optimizing physical reconfiguration parameters and control policies, as different stiffness configurations directly impact the manipulator’s kinematic behavior. 

Future work will focus on extending this approach to more complex manipulation tasks. Additionally, we aim to validate the learned policies and stiffnesses on physical prototypes to assess their real-world feasibility and robustness.

\bibliographystyle{IEEEtran}
\bibliography{IEEEabrv,reconfigurable}

\end{document}